\documentclass{article}

\usepackage{arxiv}

\usepackage[utf8]{inputenc} % allow utf-8 input
\usepackage[T1]{fontenc}    % use 8-bit T1 fonts
\usepackage{hyperref}       % hyperlinks
\usepackage{url}            % simple URL typesetting
\usepackage{booktabs}       % professional-quality tables
\usepackage{amsfonts}       % blackboard math symbols
\usepackage{nicefrac}       % compact symbols for 1/2, etc.
\usepackage{microtype}      % microtypography
\usepackage{graphicx}
\usepackage{natbib}
\usepackage{doi}
\usepackage{hyperref}
\usepackage{xcolor}
\definecolor{darkblue}{rgb}{0, 0, 0.5}
\hypersetup{colorlinks=true,citecolor=darkblue, linkcolor=darkblue, urlcolor=darkblue}

\usepackage{algorithm}
\usepackage{algorithmicx}
\usepackage[noend]{algpseudocode}
\algdef{SE}[DOWHILE]{DoWhile}{EndDoWhile}{\algorithmicdo}[1]{\algorithmicwhile\ #1}%

\usepackage{enumitem} % to change the enumerate numbering format
\usepackage{amsmath} % for equation environment 
\usepackage{amssymb} % for additional math symbols, like \R
\usepackage{mathrsfs} % for script letters in math mode
\usepackage{mathtools} % for \mathrlap
\usepackage{graphicx}
\usepackage{latexsym}
\usepackage{url}
\usepackage{xcolor}
\usepackage[normalem]{ulem}

\usepackage{amsthm}
\theoremstyle{definition}
\usepackage{linguex} % for linguistic examples
\usepackage{xspace} % for \xspace command

\usepackage{tcolorbox} %

\usepackage{pmboxdraw} % for CPC figure
\usepackage{cognac}
\newcommand{\codeurl}{\footnote{The code and data are not currently available in a public repository. Interested researchers can contact the authors to request access. We plan to make the code and data publicly available in a common repository in the near future.}}

\newcommand{\ul}[1]{\underline{#1}}
\newcommand{\cpc}[1]{\texttt{#1}}
\newcommand{\model}[1]{\texttt{#1}}
\newcommand{\Patentify}{\textit{Patentify}\xspace}
\newcommand{\tfidf}{\textit{tf-idf}\xspace}

\title{Automated Neural Patent Landscaping in the Small Data Regime}

\date{} 					% Or removing it

\author{Tisa Islam Erana \& Mark A. Finlayson \\
        Florida International University \\
        Knight Foundation School of Computing and Information Sciences \\  
        11200 S.W. 8th Street, Miami, FL 33199 USA \\
        \texttt{\{tisla016, markaf\}@fiu.edu} 
}

\renewcommand{\headeright}{Islam Erana and Finlayson}
\renewcommand{\undertitle}{}
\renewcommand{\shorttitle}{Automated Neural Patent Landscaping}

\hypersetup{
pdftitle={Automated Neural Patent Landscaping in the Small Data Regime},
pdfsubject={Automated Neural Patent Landscaping},
pdfauthor={Tisa Islam Erana, Mark A. Finlayson},
pdfkeywords={Patent Landscaping, Active Learning, Deep Neural Network, BERT for Patents, Citation Network, Classification Code},
}

\begin{document}
\maketitle

\begin{abstract}
Patent landscaping is the process of identifying all patents related to a particular technological area, and is important for assessing various aspects of the intellectual property context. Traditionally, constructing patent landscapes is intensely laborious and expensive, and the rapid expansion of patenting activity in recent decades has driven an increasing need for efficient and effective automated patent landscaping approaches. In particular, it is critical that we be able to construct patent landscapes using a minimal number of labeled examples, as labeling patents for a narrow technology area requires highly specialized (and hence expensive) technical knowledge. We present an automated neural patent landscaping system that demonstrates significantly improved performance on difficult examples (0.69 $F_1$ on `hard' examples, versus 0.6 for previously reported systems), and also significant improvements with much less training data (overall 0.75 $F_1$ on as few as 24 examples). Furthermore, in evaluating such automated landscaping systems, acquiring good data is challenge; we demonstrate a higher-quality training data generation procedure by merging \citeauthor{Abood2018AutomatedPL}'s (\citeyear{Abood2018AutomatedPL}) ``seed/anti-seed'' approach with active learning to collect difficult labeled examples near the decision boundary. Using this procedure we created a new dataset of labeled AI patents for training and testing. As in prior work we compare our approach with a number of baseline systems, and we release our code and data for others to build upon\codeurl.
\end{abstract}


\section{Introduction}\label{sec:Introduction}

In its simplest form, patent landscaping is the process of identifying all patents that are related to a particular technology or technology area. Patent landscapes are useful for a number of activities: it is important for assessing the coverage, value, or context of particular pieces of intellectual property, or for understanding the direction, speed, or concentration of innovation in a particular industry~\cite{hunt2007patentsearching}. For example, companies create patent landscapes to evaluate the risks posed by competitors in a particular technology space, or to decide whether and how much to invest in pursuing particular innovations. Patent offices and economic monitoring organizations use patent landscapes to evaluate how a particular technology is affecting or might affect the economy, for example, how much economic investment is underway in a technology, how much economic value has been generated, or how many industries or companies are supported by a particular technology. Governments, in turn, can use that information to implement technology policies, for example, deciding whether to steer investment or tax incentives to companies working in particular areas (e.g., AI or green technologies). While the simplest form of patent landscaping merely identifies which patents are related to a particular area, other more sophisticated forms of patent landscaping can seek to identify how different subareas of a technology area are related, which companies or inventor groups are the most prolific, what regions are involved, or what specific types of innovations are the focus of current development. 

Patent landscaping must be clearly differentiated from patent {\it classification}. Patent classification refers to the assignment of a patent application or patent to one or more standardized technology classes, such classes those found in the Cooperative Patent Classification (CPC) hierarchy used by the US Patent \& Trademark Office (USPTO) and the European Patent Office (EPO). Patent classification is an early (now usually automated) step in patent examination---the process of a patent office evaluating a patent application--- and is used to route a patent application to patent examiners with the correct expertise. Patent landscaping differs from patent classification because a standardized patent class scheme may not contain classes or distinctions relevant to a particular patent landscaping question or need.

Patent landscaping, even its simplest form, can be intensely laborious and expensive for at least two reasons. First, the technical expertise needed to evaluate whether a patent should be included in or excluded from a landscape is often quite specialized, and the experts possessing that knowledge are rare and their time is expensive. Consider, for example, a company that develops airbags in vehicles, and needs to know the patent landscape related to \textit{sensor-integrated variable and adaptive ventilation, opening via pressure regulation and resulting in enhanced occupant protection}. There may only be a handful of people in the world with this expertise. Second, the number of patents and patent applications is rapidly increasing and shows no signs of abating: the number of patents issued per year by the USPTO alone has doubled since 2002, from a total of 177,312 to 361,435 patents~\cite{uspto2022workload}. Indeed, after 228 years of patenting activity, the USPTO issued it's ten-millionth patent in 2018. The eleven-millionth patent was issued a mere 3 years later. Therefore, automated approaches to patent landscaping are sorely needed. 

As will be reviewed in Section~\ref{sec:related}, at least two research systems have been developed to tackle automated patent landscaping~\cite[c.f.][]{Abood2018AutomatedPL,Choi2019DeepPL}. As is the case with much recent work in NLP, these are deep neural systems, and they presumably can be adapted to different technology areas given new labeled data. While effective in many ways, these prior systems have two shortcomings. First, due to challenges of assembling good labeled data, both of these approaches were trained only on ``easy'' examples found via keyword search or patent family expansion; as a consequence, they do not work well on ``hard'' examples near the landscape boundary. Second, the models were trained on several thousand positive examples and tens of thousands of negative examples. While this generates good performance (at least on ``easy'' examples), for many technology areas this amount of labeled data would in many cases be prohibitive, in that it would extremely expensive to obtain via manual labeling, and many technology areas (especially if quite specialized) might not even contain that many positive examples. Further, such a training setup also presents data balance issues.

In this work we attempt to address these two issues (data quality and amount of data needed). First, starting from the ``seed/anti-seed'' approach of \citeauthor{Abood2018AutomatedPL} which generates ''easy'' examples, we leverage active learning to collect a set of labeled ``hard'' examples (positive and negative) near an estimated landscape boundary. This allows us to build a much higher quality dataset for training and testing and evaluate prior architectures as to their performance on these more difficult cases. Second, we investigate the use of additional features---in particular citation networks (which have not been previously used in conjunction with deep neural methods), and CPC code embeddings \cite[as was done in][]{Pujari2022PL}---which significantly improve performance when using a small number of examples. We show an overall performance of 0.75 $F_1$ on as few as 24 examples, which is comparable to prior work trained on hundreds of times as much data. 

The paper is organized as follows. We first discuss the prior work on automated patent landscaping (\S\ref{sec:related}). Next we present our data collection approach, reviewing our annotation tool and the dataset we generated using that tool (\S\ref{sec:data}). We then describe our architecture, which builds on a variety of prior work but combines them in new ways (\S\ref{sec:methods}). We review our experiments and results next (\S\ref{sec:results}. Finally, we conclude with a brief discussion (\S\ref{sec:discussion}) and a list of our contributions (\S\ref{sec:contributions}). 

\section{Related Work}
\label{sec:related}

\subsection{Patent Classification}
As noted above, patent classification (as opposed to patent landscaping) is the process of assigning a patent classification code from an established code list (such as the CPC code hierarchy) to aid patent examination and search. An example of a section of the CPC code hierarchy is shown in Figure~\ref{fig:cpccodes}. Patent classification is related to, but must be kept distinct from, patent landscaping. Traditional approaches to patent classification involve converting text features of a patent---such as abstract, claim or title---into feature vectors, and then passing these features through a supervised classification model, such as a Support Vector Machine or a K-Nearest Neighbour model \cite{YUN2020PC_TopicModelling,Seneviratne2015PC_KNN}. Other work has used the bibliographic metadata as the primary feature for labelling patent with classification codes. For example, \citet{Li2007AutomaticPC} used the CPC codes of the patents cited by a patent---both single hop (direct citation) and multi-hop (citation networks)---to create features for classification in a featurized supervised machine learning model. More recent approaches use deep neural models for patent classification. Since patents contain different types of text data like titles, abstracts, claims etc. embedding techniques like Word2Vec~\cite{Mikolov2013Word2Vec} and language models like BERT~\cite{Devlin2018BERT} have been used to great effect. For example, \citet{Lee2020PatentCB} fine-tuned BERT using a patent classification task, resulting in a model they call PatentBERT. DeepPatent~\cite{Li2018DeepPatentPC} used a Convolutional Neural Network and word embeddings of the patent text to classify patents into CPC codes.

\begin{figure*}
\begin{tcolorbox}
\footnotesize
%\color{red}%
\begin{verbatim}
{CPC Section: A - HUMAN NECESSITIES
  └ CPC Class: A01 - AGRICULTURE; FORESTRY; ANIMAL HUSBANDRY; HUNTING; TRAPPING; FISHING
    └ CPC Subclass: A01B - SOIL WORKING IN AGRICULTURE OR FORESTRY; PARTS, DETAILS, OR  
      │                    ACCESSORIES OF AGRICULTURAL MACHINES OR IMPLEMENTS, IN GENERAL
      └ CPC Group: A01B1/00 - Hand tools
        └ Main Group: A01B1/02 - Spades; Shovels 
          └ Subgroup: A01B1/024 - Foot protectors attached to the blade}
\end{verbatim}
\end{tcolorbox}
\caption{A section of the CPC code hierarchy showing CPC class, subclass, main group and subgroups}
\label{fig:cpccodes}
\end{figure*}

\subsection{Automated Patent Landscaping}

There have been several lines of effort that directly addressed the problem of automated patent landscaping. \citet{Abood2018AutomatedPL} presented a deep learning based method to select patents relevant to a particular topic from a broader set of candidate patents. Since there was no benchmark dataset available for training patent landscaping models, they developed a general method to build training data for any particular topic. Their method begins with taking a small number (~1000) of ``seed'' patents curated by experts that are the positive examples of a technology. From these they created a ``Level 1'' (L1) set by finding patents that share CPC codes with or are cited by the seed patents. A ``Level 2'' (L2) set is created by including patents that share a ``family'' relationship with those in L1 (an expression of patent priority or patent continuation). Any patent outside of L2 is considered a negative example (or an ``anti-seed'') for the topic. They then trained a deep neural network using seeds and anti-seeds as positive and negative training examples, and then applied the classifier to the patents in L2 to find patents that should be included in the landscape (which already includes the seeds). Their model architecture was a wide and deep Long-Short Term Memory~\cite[LSTM;][]{Hochreiter1997LSTM}, and they used Word2Vec~\cite{Mikolov2013Word2Vec} embeddings of words in the abstract as the sequential input input to the LSTM network. The metadata of the patents---the citations and the CPC codes---were represented by 1-hot encoding vectors. The metadata vectors and the output of the LSTM handling the abstract were concatenated and used as input to a multi-layer perceptron (MLP) network to perform a binary classification. They built landscaping models for four topics: \textit{browser}, \textit{operating system}, \textit{video codec} and \textit{machine learning}. They reported $F_1$ scores above 0.95 for each of these topics. Despite this impressive performance, the work has several shortcomings. First, the training data generated by the seed/anti-seed method does not provide positive or negative examples near the landscape boundary (i.e., they only include ``easy'' examples). In our experiments we obtain patents near the landscape boundary via human labeling and show that \cite{Abood2018AutomatedPL}'s performance is quite low on these examples. A smaller problem is that if a patent that does not share a CPC code, citation, or family link with one of the seed patents, it will never be included in the landscape (because the approach \textit{filters} patents from the L1 and L2 sets), which is potentially problematic for broad landscapes covering many CPC codes, or landscapes that include many patents (where citation networks are unlikely to be exhaustive). \citeauthor{Abood2018AutomatedPL}'s work was continued by \citet{Giczy2021IdentifyingAI},  where they removed the CPC codes and added the claim texts as features. They did not measure the model performance; but rather they used it to create a landscape of AI patents. 

\citet{Choi2019DeepPL} is the second line of effort, which also uses deep learning. They used word2vec embedding of the patent abstracts as inputs to a modified transformer architecture comprising both multi-head self-attention and scaled dot product attention. They also experimented with diffusion graph embedding techniques for representing IPC, CPC, and USPC classification codes. They used four topics analyzed in the Korea Intellectual Property Strategy Agency (KISTA) Patent trends reports~\cite{KISTA2019Patentmap}, which provided positive examples for training. They generated negative examples by repeating the keyword searches used by the human experts when generating the original landscapes, and using patents returned by that search but not included in the landscape. Because negative examples vastly outnumber positive examples, they performed under-sampling of negative examples using CPC codes of the target patents. They reported $F_1$ scores ranging from 0.62 to 0.89. They also evaluated \citeauthor{Abood2018AutomatedPL}'s approach on their data, and in all cases showed better performance. They also compared with a baseline model based on PatentBERT which performed similarly to their model.

\citet{Antonin2023Identifying} also investigated neural approaches to landscaping, using a modified version of Abood \& Feltenberger's data collection procedure. They first defined rules to identify a set of manually pre-identified candidates: these rules leveraged features such as technological classes, keywords, and patent similarity. These rules were then used to define the initial seed set, which was manually reviewed. Then they performed seed expansion as described in Abood \& Feltenberger. They experimented with different neural models, including multi-layer perceptron (MLP), convolutional neural nets (CNN), and a transformer architecture. They reported $F_1$ scores ranging from 0.65 to 0.97, although for nearly all experiments their techniques do not outperform \citeauthor{Abood2018AutomatedPL}'s approach.

Finally, \citet{Pujari2022PL} developed systems for exposing structure \textit{inside} of a landscape (which they called ``patent-landscape-oriented target classification''). In their task, they started with a technology area and had experts provide a set of categories that represented important features or aspects of the patents in the landscape, and labeled a part of landscape with these categories (which was a multi-label classification task) They experimented on three datasets---focused on the topics of \textit{Injection Values}, \textit{Ritonavir}, and \textit{Atazanavir}---and used the same classification architecture they proposed in previous work on patent classification~\cite{Pujari2021AMA}. This architecture comprised a Transformer-based Multi-task Model (TMM) which took as input SciBERT~\cite{Beltagy2019SciBERT} embeddings for title, abstract, claims, and description. They also experimented with different ways of computing graph embeddings of the CPC labels, which were used as an additional input. They achieved $F_1$s of 0.68 to 0.84.

\section{Data}
\label{sec:data}

As discussed above, the data collection method developed by \citeauthor{Abood2018AutomatedPL}, and used in follow-on papers by \citeauthor{Giczy2021IdentifyingAI} and \citeauthor{Antonin2023Identifying}, starts from a set of manually annotated positive examples (called ``seeds'', of which they used around 1,000 seeds for each  technology domain). They are able to rapidly generate a large number of negative examples (``anti-seeds'') by sampling from patents which do not share a CPC code with a seed, do not have a patent family relationship with the seeds, and are not cited by the seeds or their family members. While this method is highly scalable, we had questions about whether these examples were actually modeling the landscape boundary well. It seemed plausible that, especially for a large technology domain, the seeds would not necessarily be near the boundary, and anti-seeds would almost certainly be far from it. 

To investigate this hypothesis, we sought to collect manually annotated positive and negative examples that were near the landscape boundary, to see whether \citeauthor{Abood2018AutomatedPL}'s technique maintained it's performance in this region. The idea was to start from seed/anti-seeds collected in the manner described by \citeauthor{Abood2018AutomatedPL}, but then use these to identify examples nearer to the landscape boundary and manually annotate them. We used active learning and an annotation tool of our own design to accomplish this, as described later in this section.

To begin, we chose the technology domain of \textit{Artificial Intelligence} (AI) as the defining topic of the landscape, and worked exclusively on patents downloadable electronically from USPTO's \textit{PatentsView} \cite{uspto2020patentsview} data repository. This set numbered more than 2 million patents as of 2021. We further obtained 2,020 seed positive examples of patents in the AI space from the USPTO itself, as reported by \citet{Giczy2021IdentifyingAI}. These examples were annotated by patent examiners with expertise in AI employed by the USPTO. The USPTO also had performed \citeauthor{Abood2018AutomatedPL}'s L1/L2 expansion to generate a set of anti-seeds / negative examples, from which they randomly sampled 56,093 anti-seeds which they provided to us.

\begin{figure*}
\includegraphics[width=\textwidth]{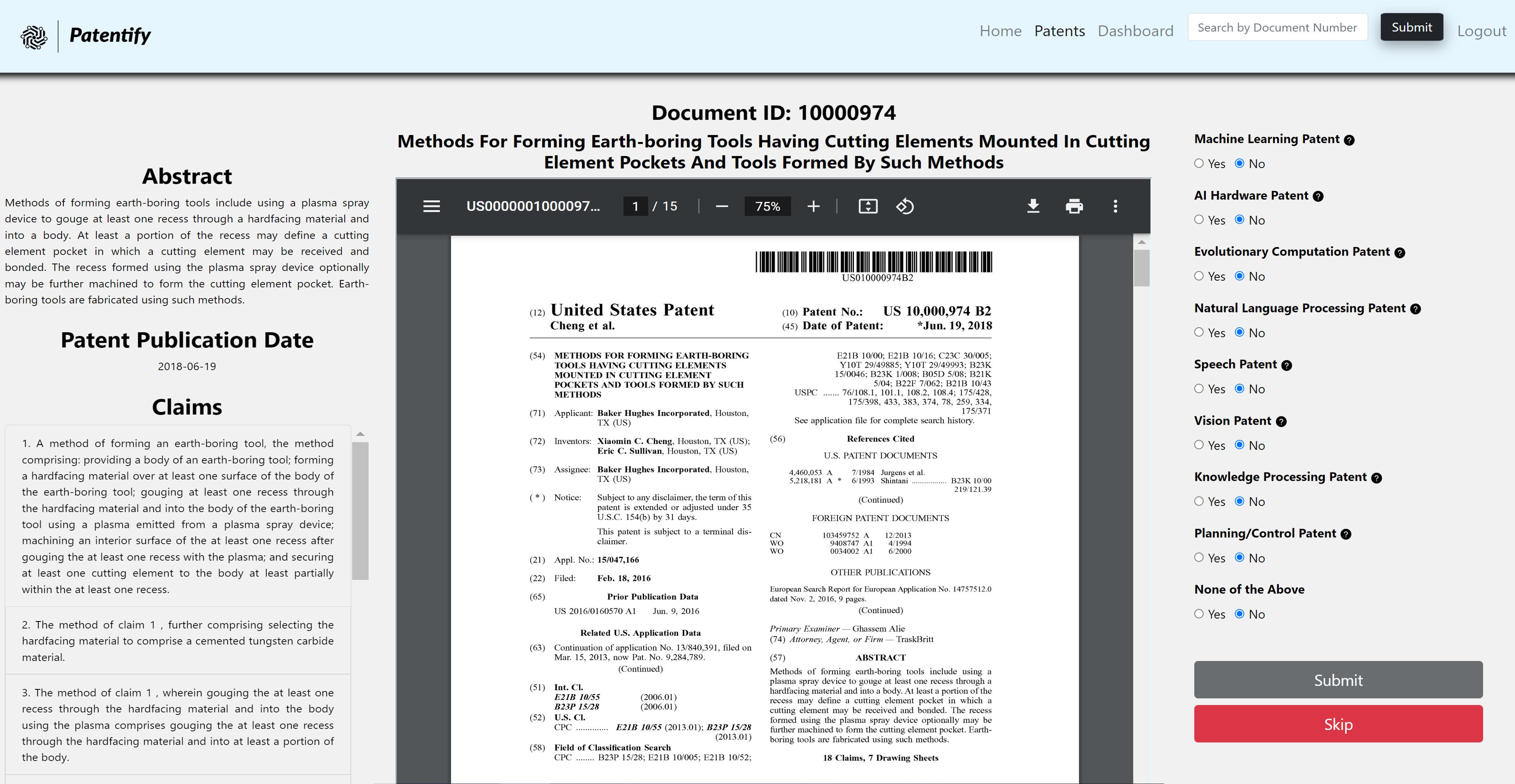}
\caption{Screenshot of the \Patentify Annotation Tool}
\label{fig:patentify}
\end{figure*}

\subsection{Annotation using Active Learning}

\textit{\Patentify}, our annotation tool, has a front-end user interface for labeling patents (shown in Figure~\ref{fig:patentify}), and runs an active-learning-based Support Vector Machine~\cite[SVM;][]{Boser1992MarginClassifier} on the back-end to identify patents near the decision boundary for labeling. Active learning is a semi-automatic machine learning technique which interacts with the user in a feedback loop and results in higher quality and more informative labeled examples~\cite{settles2019active}.

The SVM model uses as features the \tfidf counts of all the words (except stop words) in the title and abstract text of patents. We generated the initial state of model by providing it a balanced set of 938 training examples, equally balanced between four groups:  randomly sampled seeds and anti-seeds from the USPTO-provided data, and a small set of positive and negative examples near the decision boundary manually annotated by two researchers in our lab.  This model was then run over all 2 million remaining patents from the PatentsView data. We then used uncertainty sampling~\cite{Lewis1994TextClassifier} to prioritize the patents that the model is most uncertain about. These decision boundary example patents are then presented to the user for labelling.

Our annotators were graduate and undergraduate students in our laboratory with at least 2 years of research experience in AI. Each student was asked to annotate approximately 100 patents as whether or not they involved AI. After every 10 labeled examples (positive or negative) collected, \Patentify retrained the SVM model and reranked the most informative patents for labeling. Individually annotators spent approximately four hours each to do their annotations, for a total annotation effort of around fifty hours. Interannotator agreement was measured at 0.806 Cohen's $\kappa$, which is considered ``almost perfect''~\cite{Landis1977Kappa} or ``excellent''~\cite{Fleiss1981Stats} agreement.

\subsection{Dataset}\label{sec:dataset}

We collected 1,149 annotated examples comprising 395 positive examples and 754 negative examples. We then separated our data into four categories: positive vs. negative examples crossed hard vs. easy examples, as shown in Table~\ref{tab:data}.

\begin{table}[t]
\footnotesize
\centering
\begin{tabular}{@{}l@{~~}l@{~~}l@{}}
\toprule
\bf Type & \bf \# & \bf How collected \\
\midrule
Easy+ (seeds)      & 2,020      & Manual anno. by USPTO exam. \\
Easy- (anti-seeds) & 56,093     & Samples outside of L1/L2 \\
Hard+              & 395        & Manual anno. by AI students \\
Hard-              & 754        & Manual anno. by AI students \\
\bottomrule
\end{tabular}
\caption{Data collected for each type of example (hard vs. easy $\times$ positive vs. negative).}
\label{tab:data}
\end{table}

\begin{table}[t]
\footnotesize
\centering
\begin{tabular}{@{}l@{~~}l@{~}l@{~}l@{~}l@{~~}l@{}}
\toprule
\bf Dataset             & \bf Hard+ & \bf Hard- & \bf Easy+  & \bf Easy-   & \bf Total \\
\midrule
All Data / Unbalanced   & 395       & 754       & 2,020      & 56,093      & 59,262 \\
Full Balanced           & 395       & \ul{395}  & \ul{395}   & \ul{395}    & \ul{1,580} \\
Full Holdout Test       & 0         & \ul{359}  & \ul{1,625} & \ul{55,698} & \ul{57,682} \\
\bottomrule
\end{tabular}
\caption{Summary of the datasets. Underlined sets are randomly sampled and vary when generating different versions to investigate statistical variation.}
\label{tab:datasets}
\end{table}

\begin{figure*}[t]
\centering
\includegraphics[width=0.8\textwidth]{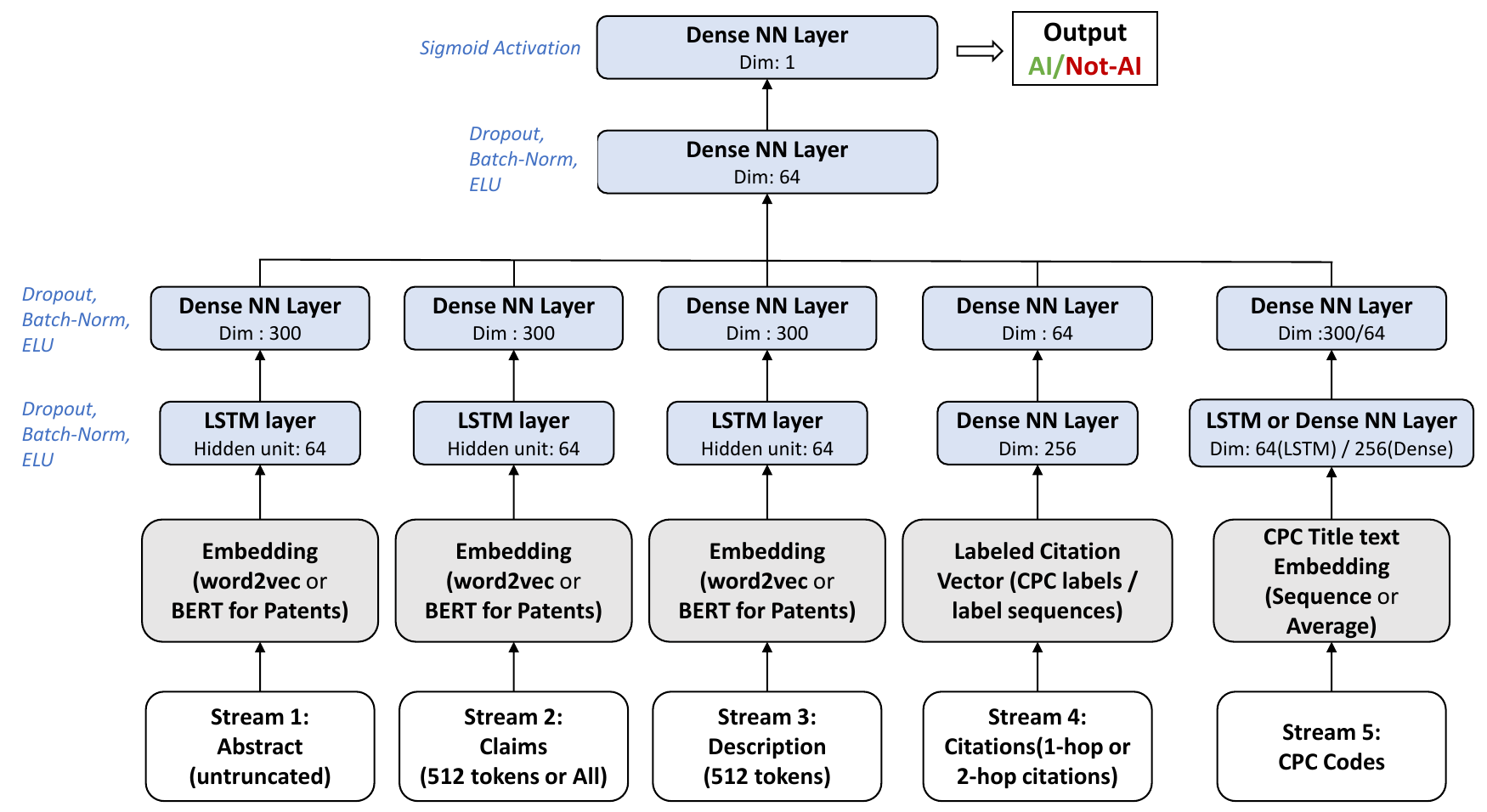}
\caption{Neural Architecture. We experimented with a number of variants. In particular, model variants included or excluded various streams (numbered 1--5) with various settings, as described in Table~\ref{tab:fullmodels}.}
\label{fig:arch}
\end{figure*}

These data allowed us to construct several different types of training and testing datasets, as shown in Table~\ref{tab:datasets}. First, we constructed an \textit{All Data / Unbalanced} set with 59,262 examples (2,415 positive, 56,847 negatives), comprising all the data. Second, we constructed a \textit{Full Balanced} dataset comprising all of the hard positive examples, and an equal number of examples randomly sampled from the other categories, resulting in 1,580 total examples (790 positive, 790 negative). In some experiments below we generated multiple versions of the Full Balanced dataset to investigate statistical variation of the data. Third, we constructed a \textit{Full Holdout Test} dataset from the examples not included in the Full Balanced set, comprising 57,682 examples (359 `hard' negatives, 1,625 'easy' positives, and 55,698 'easy' negatives). Again, when we generated a different Full Balanced dataset, this necessarily generated a complementary Full Holdout Test dataset.

\section{Patent Landscaping Methods}
\label{sec:methods}

Our approach begins from the architecture described by \citet{Abood2018AutomatedPL}, and we show how we were able to achieve higher performance both using our higher quality training data and different embedding strategies. Patents are structured documents and contain several different types of information that can be used to inform landscaping, with the two major types of information being text and metadata. Text information comprises the actual language of the patent and includes fields such as the \textit{title}, \textit{abstract}, \textit{claims}, and \textit{description} (which can be further subdivided in some cases).  Of the text information, the abstract provides a brief summary of the invention, the description provides expanded explanation and context for the invention, and the claims describe the exact legal scope of the invention.  Metadata includes classification codes, citations, names of inventor, assignee and applicant, filing and publication dates, and so forth. The classification codes and citations provide the most information for the purposes of basic landscaping.

\subsection{Our Neural Architecture}

\citeauthor{Abood2018AutomatedPL}'s model used the abstract, citations, and CPC codes. Each word in the abstract was encoded with word2vec and then fed into an LSTM followed by a dense MLP. Citations were encoded as a 1-hot vector then fed through two dense MLP layers. CPC codes were also encoded as a 1-hot vector and fed through two dense MLP layers. All three streams were fed into a single dense MLP followed by a dense binary classification layer. \citet{Giczy2021IdentifyingAI} refined \cite{Abood2018AutomatedPL}'s implementation by removing the CPC codes and adding the claim text. Our model (Figure~\ref{fig:arch}) builds on these two approaches. We have five input streams: three for text, one for citations, and one for CPC codes (they are numbered for ease of reference in the figure). As shown, text is encoded using embeddings (second layer of boxes from the bottom) and then fed into LSTM layers (third layer). Citation information and the average CPC code embedding vectors (second layer) are fed into single dense layers (third layer). All streams end with a single dense layer (fourth layer), which then combine in single dense layer (fifth layer) before passing into a binary classification layer (sixth layer).

\begin{table}[t]
\footnotesize
\centering
\begin{tabular}{@{}ll@{~~}l@{~~}l@{~~}l@{~~}l@{}}
\toprule
\bf Stream             &          & \bf 2      & \bf 3    & \bf 4       & \bf 5 \\
                       & \bf Word & \bf Claims & \bf Desc.& \bf Cite    & \bf CPC \\
\bf Model              & \bf Emb. & \bf Len.   & \bf Len. & \bf \# Hops & \bf Emb. \\
\midrule
\model{A\&F}           & w2v      & -          & -        & 1-Hot   & 1-Hot \\
\model{A\&F/USPTO}     & w2v      & Full       & -        & 1-Hot   & - \\
\midrule
\model{w2v-FullClaims} & w2v      & Full       & -        & -       & - \\ % no learning curve
% \model{B4P-FullClaims} & B4P      & Full       & -        & -       & - \\
\model{512ClaimsOnly}  & B4P      & 512        & -        & -       & - \\
\model{Plus512Desc}    & B4P      & 512        & 512      & -       & - \\
\model{1Hop}           & B4P      & 512        & -        & 1       & - \\
\model{2Hop}           & B4P      & 512        & -        & 2       & - \\ 
\model{CPCSeq}         & B4P      & 512        & -        & -       & Seq. \\
\model{CPCAvg}         & B4P      & 512        & -        & -       & Avg. \\
\model{B4P+All}        & B4P      & 512        & 512      & 1       & Avg. \\
\bottomrule
\end{tabular}
\caption{List of model variants, showing which embedding is used and which streams they incorporate with which settings. All models used stream 1 (abstracts). w2v = word2vec, B4P = Bert for Patents.}
\label{tab:fullmodels}
\end{table}

\paragraph{Textual Information (Streams 1--3)}
For text we experimented with using the abstract, claims, and description, encoded with word2Vec \cite{Mikolov2013Word2Vec} or BERT for Patents~\citet{Srebrovic2020BERTPatents}. BERT for Patents was pre-trained on 100 million patents and patent-related documents, with an input width of 512 tokens. It also provides a patent-specific tokenizer. We extracted the contextualized token embeddings for abstracts and claims from the second to last encoder layer. While 512 tokens is too small for most claims sections, one can chunk the text into 512-token-sized pieces and pass them through the model one at a time to create an embedding vector longer than 512 that encodes an arbitrarily long text. When using the BERT for Patents  on abstracts and claims sections, we experimented with using a single 512 token chunk (single pass) as well as using all of the text (multiple passes). We only used a single pass for descriptions because of their length. Note that Patents are public documents, part of the public record, and contain the names and addresses of inventors and assignees (which is personally identifiable information). We did not use this information in training the models.

\begin{table*}[t]
\footnotesize
\centering
\begin{center}
\begin{tabular}{ll|lll|lll|lll}
\toprule
           &  &  \multicolumn{3}{c|}{\bf Hard} & \multicolumn{3}{c|}{\bf Easy} & \multicolumn{3}{c}{\bf Holdout} \\
\bf Models & \bf Overall & \bf Avg. & \bf + & \bf - & \bf Avg. & \bf + & \bf - & \bf Hard- & \bf Easy+   & \bf Easy- \\

\midrule
% Baselines
% Overall Easy Hard % 
\model{A\&F}           & 0.68  & 0.60  & 0.62  & 0.57  & 0.77  & 0.80  & 0.74  & 0.39  & 0.95 & 0.43 \\ 
\model{A\&F/USPTO}     & 0.73  & 0.62  & 0.63  & 0.60  & 0.82  & 0.83 & 0.82  & 0.38  & 0.95 & 0.79   \\
\model{Choi}           & 0.77  & 0.63  & 0.63  & 0.63  & 0.92  & 0.91 & 0.92  & 0.42  & 0.95 & 0.76   \\
\model{SVM/1Hop}       & 0.77 & 0.65  & 0.60 & 0.69  & 0.89  & 0.89 & 0.89  & 0.89  & 0.94  & 0.93  \\    
\model{SVM/tfidf-1Hop} & 0.77 & 0.66  & 0.61  & 0.69  & 0.89  & 0.89 & 0.89  & 0.88  & 0.95 & 0.93  \\  
\model{SVM/tfidf}      & 0.78  & 0.67  & 0.64  & 0.69  & 0.89  & 0.89 & 0.88  & 0.84  & 0.94 & 0.9   \\  
\model{SVM/w2v}        & 0.73 & 0.67 & 0.64  & 0.69 & 0.79 & 0.78 & 0.81  & 0.55  & 0.69 & 0.71  \\
\model{SVM/FT}         & 0.78 & 0.68 & 0.64  & 0.68 & 0.88 & 0.85 & 0.91  & 0.84  & 0.93 & 0.90  \\
% \model{SVM/BERT-CLS}   & 0.14 & 0.11 & & & 0.16  \\

\midrule
%Neural
%&  &  &  &  &  & &  &  & &
\model{w2v-FullClaims} & 0.68 & 0.59 & 0.54 & 0.64 & 0.78 & 0.75 & 0.81 & 0.86 & 0.81 & 0.90  \\
\model{2Hop}           & 0.76 & 0.62 & 0.57 & 0.67 & 0.90 & 0.90  & 0.90  & 0.82 & 0.96 & 0.88 \\
\model{CPCSeq}         & 0.76 & 0.64 & 0.65 & 0.63 & 0.88 & 0.89 & 0.86 & 0.79 & 0.98 & 0.88   \\
\model{512ClaimsOnly}  & 0.77 & 0.66 & 0.64 & 0.68 & 0.88 & 0.89 & 0.86 & 0.81 & 0.98 & 0.89   \\
\model{Plus512Desc}    & 0.77 & 0.65 & 0.63 & 0.66 & 0.90  & 0.90  & 0.89 & 0.84 & 0.98 & 0.9   \\
\model{1Hop}           & 0.77 & 0.63 & 0.60 & 0.66 & 0.90  & 0.90  & 0.90  & 0.85 & 0.98 & 0.91   \\
\model{CPCAvg}         & \bf 0.80 & \bf 0.69 & \bf 0.67 & \bf 0.71 & 0.91 & 0.92 & 0.91 & 0.82 & 0.99 & 0.92   \\
\model{B4P+All}        & 0.79 & 0.64 & 0.59 & 0.70 & \bf 0.93  & \bf 0.93 & \bf 0.93 & \bf 0.89 & \bf 0.99 & \bf 0.95  \\
\bottomrule
\end{tabular}
\end{center}
\caption{Average 5-fold cross-validated $F_1$ scores of all models. The \textbf{Overall} is the average between \textbf{Hard} and \textbf{Easy}, excluding the \textbf{Holdout} results. Best scores for each column are boldfaced. Standard deviations were 0.32 for \model{A\&F}, 0.29 for \model{w2v-FullClaims}, and no more than 0.1 for all other models.}
\label{tab:f1scores}
\end{table*}

\paragraph{Citation Networks (Stream 4)}
As shown by \citet{Li2007AutomaticPC}, citation information can be used in several ways. First, one can examine the direct outward citations of patents or patent pre-grant publications (PGPubs); this is called the \textit{direct citation} approach. Second, one can collect further citations by following citations of citations, up to a certain number of hops (\textit{citation network} approach). We experimented with both (stream 4). Direct citations is a special case of the citation network approach, with hops is limited to 1. \citeauthor{Abood2018AutomatedPL} used the direct citation (1-hop) approach, but encoded citations in a 1-hot vector, which is problematic because of the sparseness. In our experiments with \citeauthor{Abood2018AutomatedPL}'s architecture, we found that removing the 1-hot citation information actually improved performance. In our approach, we encoded 1-hop citations by representing each document as a vector of counts of CPC codes at the subclass level. Each element of the input vector to the first dense layer thus contains the number of cited documents belonging to a particular CPC subgroup code. We also experimented with using CPC codes gathered from citations two hops away from the target patent (namely, a 2-hop citation network). These were encoded as a vector of counts of two-code sequences; i.e., if the the first patent in a two-hop citation chain had code \cpc{A01B}, and the second patent had code \cpc{E05D}, then the sequence \cpc{A01B-E05D} would be incremented by 1. 

\paragraph{CPC Embeddings (Stream 5)}
We embedded tokens in the concatenated CPC titles using BERT for Patents. We tried two approaches. First, we fed the embedding for each token, for each code in sequence, one at a time into an LSTM layer. Second, we computed an average embedding for the set of codes for a patent by averaging the embeddings of the tokens, place-wise, meaning the embedding vectors for token 1 of all codes was averaged together to produce an average token 1 embedding, the same for token 2, and so forth. This vector was then fed into a single dense layer.

\subsection{Models Tested}

Figure~\ref{fig:arch} illustrates the design choices in our architecture. We experimented with different combinations, listed in Table~\ref{tab:fullmodels}. To compare with our models on the same data, we obtained or created implementations of \citeauthor{Abood2018AutomatedPL}'s, \citeauthor{Giczy2021IdentifyingAI}'s, and \citeauthor{Choi2019DeepPL}'s architectures. We refer to these models as \model{A\&F},  \model{A\&F/USPTO}, and \model{Choi} respectively. We also created five baseline SVM models (RBF kernel). The first baseline was a regular Bag-of-Words approach over abstracts and claims with the words represented as \tfidf vectors~\cite{Zhang2011TFIDF} (vocabulary size of 49,639 [abstracts] and 77,989 [claims]; cut-off value of 1.0). We call this baseline \model{SVM/tfidf}. The second baseline used word2vec vector embeddings of the abstracts and claims, on which we performed principal component analysis \cite[PCA]{Jolliffe1986PCA}, feeding the top 50 components as inputs to the SVM (\model{SVM/w2v}). The third baseline used FastText embeddings~\cite{Bojanowski2017Subword} in the same way as the word2vec embeddings (\model{SVM/FT}).  The fourth baseline used the vector of counts of CPC subclass appearances in direct citations of a patent as the SVM features (\model{SVM/1Hop}). The fifth baseline used a combination of the \tfidf features from the first model and the citation features from the fourth model (\model{SVM/tfidf-1Hop}). We tested 8 variants of our architecture, listed at the bottom of Table~\ref{tab:fullmodels}. That table shows which embedding was used (w2v or B4P), and which streams were used with what parameters (all models used stream 1 and 2). For example, \model{w2v-FullClaims} uses word2vec embeddings throughout, with the full claims for stream 2. On the other hand, \model{1Hop} used stream 2 with 512 tokens, and incorporated stream 4 with 1-hop citations, and B4P embeddings throughout.

\section{Results}
\label{sec:results}

\begin{figure*}[t]
    \centering
    \includegraphics[width=0.75\textwidth]{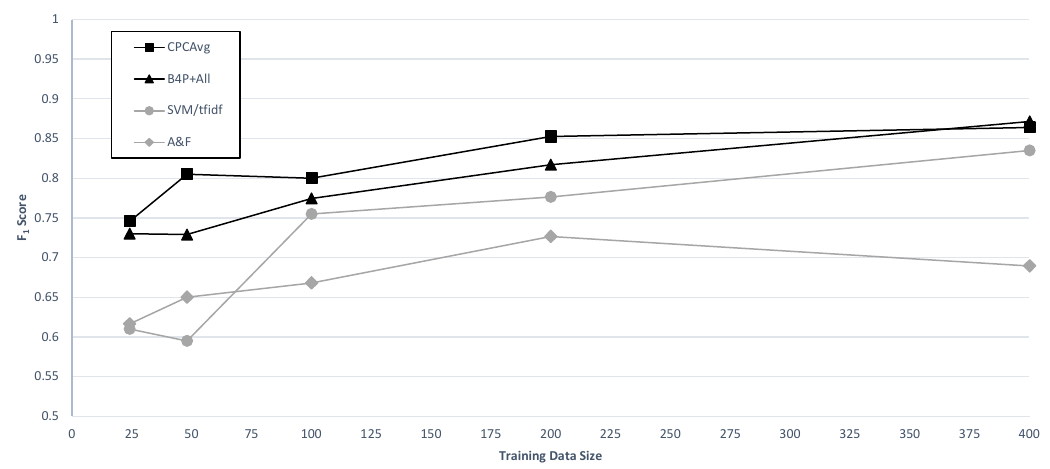}
    \caption{Learning curve for 400, 200, 100, 48, and 24 examples on the two best performing models (\model{CPCAvg} and \model{B4P-All}) compared with \citeauthor{Abood2018AutomatedPL}'s model (\model{A\&F}) and the best performing baseline (\model{SVM/tfidf}).}
    \label{fig:learningcurve}
\end{figure*}

\begin{table*}[t]
\centering
\begin{tabular}{l|l|l|l|l|l}
\toprule
\bf Models             & \bf 400   & \bf 200  & \bf 100  & \bf 48   & \bf 24 \\

\midrule
% Baselines
\model{A\&F}           & 0.69      & 0.73     & 0.67     & 0.65     & 0.62  \\ 
\model{A\&F/USPTO}     & 0.75      & 0.73     & 0.68     & 0.66     & 0.64   \\
\model{Choi}           & 0.75      & 0.68     & 0.48     & 0.48     & 0.44   \\
\model{SVM/1Hop}       & 0.83      & 0.80     & 0.73     & 0.65     & 0.61   \\    
\model{SVM/tfidf-1Hop} & 0.83      & 0.80     & 0.74     & 0.62     & 0.64   \\  
\model{SVM/tfidf}      & 0.84      & 0.78     & 0.76     & 0.60     & 0.61   \\  
\model{SVM/w2v}        & 0.69      & 0.60     & 0.59     & 0.57     & 0.53   \\
\model{SVM/FT}         & 0.67      & 0.61     & 0.58     & 0.54     & 0.53   \\

\midrule
%Neural
%&  &  &  &  &  & &
\model{B4P-FullClaims} & 0.85      & 0.82     & \bf 0.82 & 0.80     & 0.72   \\
\model{2Hop}           & 0.84      & 0.81     & \bf 0.82 & 0.75     & 0.68   \\
\model{CPCSeq}         & 0.85      & \bf 0.86 & 0.78     & 0.78     & 0.73   \\
\model{512ClaimsOnly}  & 0.86      & 0.82     & 0.78     & 0.74     & 0.73   \\
\model{Plus512Desc}    & 0.84      & 0.83     & 0.72     & 0.76     & 0.71   \\
\model{1Hop}           & 0.84      & 0.85     & 0.79     & 0.77     & 0.72   \\
\model{CPCAvg}         & 0.86      & 0.85     & 0.80     & \bf 0.81 & \bf 0.75 \\
\model{B4P+All}        & \bf 0.87  & 0.82     & 0.77     & 0.73     & 0.73   \\
\bottomrule
\end{tabular}
\caption{Overall $F_1$ scores of all models, 5-fold cross validated on 400, 200, 100, 48, and 24 training examples.}
\label{tab:LearningCurve_Scores}
\end{table*}

We trained our models for 5 epochs and in batches of 64 samples, with LSTM hidden unit size 64, dense layers having size 300, 64 and 1, both using 40\% dropout. We used the default ADAM optimizer with a learning rate of 0.0001. We performed 5-fold cross validation in all experiments. Experiments using the balanced dataset are shown in Table~\ref{tab:f1scores}. We also computed learning curves for all models using smaller balanced subsets of the balanced datasets, and the results for the best performing models are shown in Figure~\ref{fig:learningcurve}. Numbers of parameters for the models ranged from 637,753 (\model{512ClaimsOnly}) to 1,386,389 (\model{B4P+All}), which does not include parameters of the pre-trained models. Training each model took approximately two hours on a dual CPU compute node (Xeon Gold 6258R, 2.7 GHz) with 1.5TB RAM and 8 Nvidia A100 40GB HBM2 PCIe 4.0 GPUs. Data processing took approximately 4 hours for each model. 

As shown in Table~\ref{tab:f1scores}, \model{A\&F}, \model{A\&F/USPTO}, and \model{Choi} did not perform nearly as well on our data as originally reported (for example, \model{A\&F} reported an $F_1$ of 0.98; our experiments resulted in 0.68). Those models also underperformed our baselines. \model{A\&F} and \model{A\&F/USPTO} had low performance both on Hard and Easy examples. One possible explanation for this is our use of balanced data. On the other hand, the SVM models performed quite well both on Easy and Hard examples when \tfidf and Word2Vec (or FastText) embeddings were used as feature vectors for abstract and claims and the models were trained on all the data. The SVM models using citation information performed less well than the other baselines, but still reasonably. Using citation information improved performance on the Hard examples. Note that \model{w2v-FullClaims} is the same as \model{A\&F} (without citation and CPC 1-hot), and shows slightly better performance, which suggests that the 1-hot encodings add noise to the network. Table~\ref{tab:f1scores} also shows the performance of the models on the holdout datasets (Table~\ref{tab:datasets}). All models except \model{A\&F}, \model{A\&F/USPTO}, and \model{Choi} perform well on the Holdout data. The best two models on the Full Balanced Dataset are \model{CPCAvg} and \model{B4P+All}, the first performing best on Hard examples and the second performing best on Easy examples. \model{B4P+All} performs the best on the Holdout data.  

Figure~\ref{fig:learningcurve} shows learning curves for our two best performing models (\model{CPCAvg} and \model{B4P-All}), the baseline with the best learning curve (\model{SVM/tfidf}), and \citeauthor{Abood2018AutomatedPL}'s model (\model{A\&F}), which was the best performing prior model. We trained all models on Balanced Datasets of sizes 400, 200, 100, 48, and 24. When data size falls below 100, we see a marked divergence between the baselines and our neural models, reaching a difference of nearly 14 points of $F_1$ between \model{CPCAvg} and \model{B4P-All}. Importantly, the neural models show less degradation as the dataset shrinks, with \model{CPCAvg} and \model{B4P-All} dropping only 14\% and 16\%, respectively, while the \model{SVM/tfidf} drops 27\% overall (\model{A\&F} does not drop very much, but it is not good to begin with).

\section{Discussion}
\label{sec:discussion}
There are several key takeaways from the experimental results. \textbf{First}, the prior models \model{A\&F}, \model{A\&F/USPTO}, and \model{Choi} didn't perform nearly as well on balanced data that includes examples near the landscape boundary. In particular, they perform relatively poorly on Hard examples, showing that examples near the landscape boundary are critical. \textbf{Second}, the BERT for Patents (B4P) embeddings were effective in improving performance of architectures in which it was substituted for other embeddings. \textbf{Third}, for large amounts of data, most of the models, including the SVM baselines (but excluding \model{A\&F}, performed quite well and were very close in performance. This suggests that, when a lot of data is available the extra complexity and computational load of the neural methods does not purchase much in the way of performance. \textbf{Fourth}, direct citations (1-hop citations) combined with CPC codes give good result on both Easy and Hard examples, but compared with the other features does not seem to add much. Interestingly, 2-hop citations have marginally poorer performance on the Hard examples. \textbf{Finally}, where the neural models really improve over the baselines is in the low-data regime. In the regime of 24 examples (evenly balanced between Hard/Easy and Positive/Negative), we see improvements of nearly 14 points of $F_1$ over baseline.  Interestingly, the \model{CPCAvg} model, which uses only abstract text, 512 tokens of the claims, and CPC information, is the best overall model. Regardless, even this best model only achieves 0.75 $F_1$ at 24 examples, which has much room for improvement.

\section{Contributions}
\label{sec:contributions}
In this paper, we have demonstrated the performance of various neural models on a patent dataset that we have built. Our work contributes to the research of automated patent landscaping models in four different aspects. First, we have shown that previously proposed ``seed/anti-seed'', while useful, importantly lacks examples near the decision boundary and using only seed/anti-seed data gives a misleading view of model performance. Second, we demonstrate an active learning augmentation to the seed/anti-seed approach which can quickly generate high-quality data near the decision boundary. Third, we conducted systematic experiments comparing the utility of different portions of the information in a patent, concluding that abstract, claims, and CPC codes (and note description or citations) provide the most power. Fourth, when a lot of data are available (1000s of examples) simple methods like SVM work just as well as the neural architectures with different features not making much difference. Fifth, we have experimented with using citation information (1-hop and 2-hop) and showed that this information does not add much performance beyond using text in the claims. Finally, we show that the neural methods significantly outperform the baselines in the small data regime (less than 100 total training examples).

\section{Limitations}
There are several limitations of this work. First, we only examine a single patent landscaping domain, that of AI. AI is a fairly broad technology area and it is not clear that the results will generalize to more specific landscape topics. Second, there may of course be other neural architecture styles that work better; we didn't exhaustively explore these choices as we felt they were beyond scope of the paper as we envisioned it. Future work should explore the space of possible architecture designs. Third, citation network information did not improve the results as much as expected, which is counter-intuitive. We suspect that there are ways to use this information that will improve performance even with the other streams of information, but we were unable to identify them. Finally, the performance of the model in the small data regime could still be improved; it remains to be seen what the true lower limit of data is required to construct a good landscape for such a broad technology domain.

\bibliographystyle{unsrtnat}
\bibliography{main}  %%% Uncomment this line and comment out the ``thebibliography'' section below to use the external .bib file (using bibtex) .

\begin{thebibliography}{28}
\providecommand{\natexlab}[1]{#1}
\providecommand{\url}[1]{\texttt{#1}}
\expandafter\ifx\csname urlstyle\endcsname\relax
  \providecommand{\doi}[1]{doi: #1}\else
  \providecommand{\doi}{doi: \begingroup \urlstyle{rm}\Url}\fi

\bibitem[Abood and Feltenberger(2018)]{Abood2018AutomatedPL}
Aaron Abood and Dave Feltenberger.
\newblock Automated patent landscaping.
\newblock \emph{Artificial Intelligence and Law}, 26:\penalty0 103--125, 2018.

\bibitem[Hunt et~al.(2007)Hunt, Nguyen, and Rodgers]{hunt2007patentsearching}
D.~Hunt, L.~Nguyen, and M.~Rodgers.
\newblock \emph{Patent Searching: Tools \& Techniques}.
\newblock Wiley, 2007.
\newblock ISBN 9780470116838.
\newblock URL \url{https://books.google.com/books?id=iXpNx4n1fDwC}.

\bibitem[USPTO(2022)]{uspto2022workload}
USPTO.
\newblock {FY} 2022 workload tables.
\newblock \url{https://www.uspto.gov/about-us/performance-and-planning/uspto-annual-reports}, 2022.
\newblock Last Accessed July 10, 2023.

\bibitem[Choi et~al.(2019)Choi, Lee, Park, and Choi]{Choi2019DeepPL}
Seok-Jai Choi, Hyeonju Lee, Eunjeong Park, and Sungchul Choi.
\newblock Deep patent landscaping model using transformer and graph embedding.
\newblock \emph{arXiv: Computation and Language}, 2019.

\bibitem[Pujari et~al.(2022)Pujari, Str{\"o}tgen, Giereth, Gertz, and Friedrich]{Pujari2022PL}
Subhash Pujari, Jannik Str{\"o}tgen, Mark Giereth, Michael Gertz, and Annemarie Friedrich.
\newblock Three real-world datasets and neural computational models for classification tasks in patent landscaping.
\newblock In \emph{Proceedings of the 2022 Conference on Empirical Methods in Natural Language Processing}, pages 11498--11513, Abu Dhabi, United Arab Emirates, December 2022. Association for Computational Linguistics.
\newblock URL \url{https://aclanthology.org/2022.emnlp-main.791}.

\bibitem[Yun and Geum(2020)]{YUN2020PC_TopicModelling}
Junghwan Yun and Youngjung Geum.
\newblock Automated classification of patents: A topic modeling approach.
\newblock \emph{Computers \& Industrial Engineering}, 147:\penalty0 106636, 2020.
\newblock ISSN 0360-8352.
\newblock \doi{https://doi.org/10.1016/j.cie.2020.106636}.
\newblock URL \url{https://www.sciencedirect.com/science/article/pii/S0360835220303703}.

\bibitem[Seneviratne et~al.(2015)Seneviratne, Geva, Zuccon, Ferraro, Chappell, and Meireles]{Seneviratne2015PC_KNN}
Dilesha Seneviratne, Shlomo Geva, Guido Zuccon, Gabriela Ferraro, Timothy Chappell, and Magali R.~G. Meireles.
\newblock A signature approach to patent classification.
\newblock In Guido Zuccon, Shlomo Geva, Hideo Joho, Falk Scholer, Aixin Sun, and Peng Zhang, editors, \emph{AIRS}, volume 9460 of \emph{Lecture Notes in Computer Science}, pages 413--419. Springer, 2015.
\newblock ISBN 978-3-319-28939-7.
\newblock URL \url{http://dblp.uni-trier.de/db/conf/airs/airs2015.html#SeneviratneGZFC15}.

\bibitem[Li et~al.(2007)Li, Chen, Zhang, and Li]{Li2007AutomaticPC}
Xin Li, Hsinchun Chen, Zhu Zhang, and Jiexun Li.
\newblock Automatic patent classification using citation network information: an experimental study in nanotechnology.
\newblock In \emph{JCDL '07}, 2007.

\bibitem[Mikolov et~al.(2013)Mikolov, Chen, Corrado, and Dean]{Mikolov2013Word2Vec}
Tomas Mikolov, Kai Chen, Greg Corrado, and Jeffrey Dean.
\newblock Efficient estimation of word representations in vector space.
\newblock \emph{CoRR}, abs/1301.3781, 2013.
\newblock URL \url{http://dblp.uni-trier.de/db/journals/corr/corr1301.html#abs-1301-3781}.

\bibitem[Devlin et~al.(2018)Devlin, Chang, Lee, and Toutanova]{Devlin2018BERT}
Jacob Devlin, Ming{-}Wei Chang, Kenton Lee, and Kristina Toutanova.
\newblock {BERT:} pre-training of deep bidirectional transformers for language understanding.
\newblock \emph{CoRR}, abs/1810.04805, 2018.
\newblock URL \url{http://arxiv.org/abs/1810.04805}.

\bibitem[Lee and Hsiang(2020)]{Lee2020PatentCB}
Jieh-Sheng Lee and Jieh Hsiang.
\newblock Patent classification by fine-tuning bert language model.
\newblock \emph{World Patent Information}, 2020.

\bibitem[Li et~al.(2018)Li, Hu, Cui, and Hu]{Li2018DeepPatentPC}
Shaobo Li, Jie Hu, Yuxin Cui, and Jianjun Hu.
\newblock Deeppatent: patent classification with convolutional neural networks and word embedding.
\newblock \emph{Scientometrics}, 117:\penalty0 721--744, 2018.

\bibitem[Hochreiter and Schmidhuber(1997)]{Hochreiter1997LSTM}
Sepp Hochreiter and Jürgen Schmidhuber.
\newblock {Long Short-Term Memory}.
\newblock \emph{Neural Computation}, 9\penalty0 (8):\penalty0 1735--1780, 11 1997.
\newblock ISSN 0899-7667.
\newblock \doi{10.1162/neco.1997.9.8.1735}.
\newblock URL \url{https://doi.org/10.1162/neco.1997.9.8.1735}.

\bibitem[Giczy et~al.(2021)Giczy, Pairolero, and Toole]{Giczy2021IdentifyingAI}
Alexander~V. Giczy, Nicholas~A. Pairolero, and Andrew~A. Toole.
\newblock Identifying artificial intelligence (ai) invention: a novel ai patent dataset.
\newblock \emph{The Journal of Technology Transfer}, 47:\penalty0 476--505, 2021.

\bibitem[KISTA(2023)]{KISTA2019Patentmap}
KISTA.
\newblock {KISTA} patent trends reports.
\newblock \url{https://biz.kista.re.kr/patentmap/front/common.do?method=main}, 2023.
\newblock Last Accessed July 14, 2023.

\bibitem[Antonin and Cyril(2023)]{Antonin2023Identifying}
Bergeaud Antonin and Verluise Cyril.
\newblock Identifying technology clusters based on automated patent landscaping.
\newblock \emph{PLOS ONE}, 18\penalty0 (12):\penalty0 1--17, 12 2023.
\newblock \doi{10.1371/journal.pone.0295587}.
\newblock URL \url{https://doi.org/10.1371/journal.pone.0295587}.

\bibitem[Pujari et~al.(2021)Pujari, Friedrich, and Strotgen]{Pujari2021AMA}
Subhash~Chandra Pujari, Annemarie Friedrich, and Jannik Strotgen.
\newblock A multi-task approach to neural multi-label hierarchical patent classification using transformers.
\newblock In \emph{European Conference on Information Retrieval}, 2021.

\bibitem[Beltagy et~al.(2019)Beltagy, Lo, and Cohan]{Beltagy2019SciBERT}
Iz~Beltagy, Kyle Lo, and Arman Cohan.
\newblock {S}ci{BERT}: A pretrained language model for scientific text.
\newblock In \emph{Proceedings of the 2019 Conference on Empirical Methods in Natural Language Processing and the 9th International Joint Conference on Natural Language Processing (EMNLP-IJCNLP)}, pages 3615--3620, Hong Kong, China, November 2019. Association for Computational Linguistics.
\newblock \doi{10.18653/v1/D19-1371}.
\newblock URL \url{https://aclanthology.org/D19-1371}.

\bibitem[{USPTO}(2019)]{uspto2020patentsview}
{USPTO}.
\newblock {Patents}{View}.
\newblock \url{https://www.patentsview.org/download/}, 2019.
\newblock Updated December 31, 2019.

\bibitem[Boser et~al.(1992)Boser, Guyon, and Vapnik]{Boser1992MarginClassifier}
Bernhard~E. Boser, Isabelle~M. Guyon, and Vladimir~N. Vapnik.
\newblock A training algorithm for optimal margin classifiers.
\newblock In \emph{Proceedings of the 5th Annual Workshop on Computational Learning Theory (COLT '92)}, COLT '92, page 144–152, New York, NY, USA, 1992. Association for Computing Machinery.
\newblock ISBN 089791497X.
\newblock \doi{10.1145/130385.130401}.
\newblock URL \url{https://doi.org/10.1145/130385.130401}.

\bibitem[Settles(2012)]{settles2019active}
Burr Settles.
\newblock \emph{Active Learning}.
\newblock Synthesis Lectures on Artificial Intelligence and Machine Learning. Morgan \& Claypool Publishers, 2012.

\bibitem[Lewis and Gale(1994)]{Lewis1994TextClassifier}
David~D. Lewis and William~A. Gale.
\newblock A sequential algorithm for training text classifiers.
\newblock In Bruce~W. Croft and C.~J. van Rijsbergen, editors, \emph{SIGIR '94}, pages 3--12, London, 1994. Springer London.
\newblock ISBN 978-1-4471-2099-5.

\bibitem[Landis and Koch(1977)]{Landis1977Kappa}
J.~Richard Landis and Gary~G. Koch.
\newblock The measurement of observer agreement for categorical data.
\newblock \emph{Biometrics}, 33\penalty0 (1):\penalty0 159--174, 1977.

\bibitem[Fleiss(1981)]{Fleiss1981Stats}
J.L. Fleiss.
\newblock \emph{Statistical methods for rates and proportions}.
\newblock John Wiley \& Sons, New York, NY, 2nd edition, 1981.

\bibitem[Rob~Srebrovic(2020)]{Srebrovic2020BERTPatents}
Jay~Yonamine Rob~Srebrovic.
\newblock Leveraging the bert algorithm for patents with tensorflow and bigquery, 2020.
\newblock URL \url{https://services.google.com/fh/files/blogs/bert_for_patents_white_paper.pdf}.

\bibitem[Zhang et~al.(2011)Zhang, Yoshida, and Tang]{Zhang2011TFIDF}
Wen Zhang, Taketoshi Yoshida, and Xijin Tang.
\newblock A comparative study of tf*idf, lsi and multi-words for text classification.
\newblock \emph{Expert Syst. Appl.}, 38\penalty0 (3), 2011.
\newblock ISSN 0957-4174.
\newblock \doi{10.1016/j.eswa.2010.08.066}.
\newblock URL \url{https://doi.org/10.1016/j.eswa.2010.08.066}.

\bibitem[Jolliffe(1986)]{Jolliffe1986PCA}
I.T. Jolliffe.
\newblock \emph{Principal Component Analysis}.
\newblock Springer Verlag, 1986.

\bibitem[Bojanowski et~al.(2017)Bojanowski, Grave, Joulin, and Mikolov]{Bojanowski2017Subword}
Piotr Bojanowski, Edouard Grave, Armand Joulin, and Tomas Mikolov.
\newblock Enriching word vectors with subword information.
\newblock \emph{Transactions of the Association for Computational Linguistics}, 5:\penalty0 135--146, 2017.
\newblock \doi{10.1162/tacl_a_00051}.
\newblock URL \url{https://aclanthology.org/Q17-1010}.

\end{thebibliography}

\end{document}